\pdfoutput=1   
\documentclass{article}
\PassOptionsToPackage{round}{natbib}
\usepackage[dblblindworkshop, final]{neurips_2026}
\workshoptitle{Grounded User Simulation for Model Evaluation and Training}

\usepackage[utf8]{inputenc}
\usepackage[T1]{fontenc}
\usepackage[hidelinks]{hyperref}
\usepackage{url}
\usepackage{booktabs}
\usepackage{siunitx}          
\usepackage{threeparttable}   
\usepackage{amsfonts}
\usepackage{amsmath}
\usepackage{nicefrac}
\usepackage{microtype}
\usepackage{graphicx}
\usepackage{float}          
\usepackage{subcaption}
\usepackage{xcolor}

\graphicspath{{figures/}}
\IfFileExists{tables/numbers.tex}{\newcommand{\nQuestions}{60}
\newcommand{\nCountries}{12}
\newcommand{\nSimulators}{6}

\newcommand{\aucD}{0.59}

\newcommand{\invalidFullPct}{55.2}
\newcommand{\invalidEightyPct}{50.9}
\newcommand{\pInvLowD}{0.456}
\newcommand{\pInvHighD}{0.673}

\newcommand{\rateDirection}{0.391}
\newcommand{\rateMean}{0.651}
\newcommand{\rateGap}{0.507}
\newcommand{\rateTV}{0.942}

\newcommand{\nCal}{7}
\newcommand{\nTest}{5}
\newcommand{\aucDirection}{0.67}
\newcommand{\aucMean}{0.51}
\newcommand{\aucGap}{0.72}
\newcommand{\aucTV}{0.47}
\newcommand{\spearmanGap}{0.37}

\newcommand{\gapInvalidFullPct}{40.8}
\newcommand{\gapInvalidEightyPct}{32.9}

\newcommand{\aucGapLo}{0.64}
\newcommand{\aucGapHi}{0.79}
\newcommand{\aucWGap}{0.50}
\newcommand{\gapAucEpsMin}{0.70}
\newcommand{\gapAucEpsMax}{0.73}
\newcommand{\gapAucNMin}{0.69}
\newcommand{\gapAucNMax}{0.76}
\newcommand{\unresolvedNlo}{22.3}
\newcommand{\unresolvedNhi}{5.4}
\newcommand{\aucGapGAS}{0.71}
\newcommand{\aucGapWVS}{0.82}

\newcommand{\invalidFrontierPct}{63.2}
\newcommand{\invalidSmallPct}{63.6}

\newcommand{\sharedFracGap}{0.58}
\newcommand{\sharedFracTV}{0.94}

\newcommand{\splitAucMed}{0.73}
\newcommand{\splitAucLo}{0.68}
\newcommand{\splitAucHi}{0.78}
\newcommand{\splitFracAbovePct}{100}
\newcommand{\nSplits}{300}
\newcommand{\locoMin}{0.62}
\newcommand{\locoMax}{0.84}
\newcommand{\locoMinCountry}{Lebanon}
\newcommand{\pairedDiff}{0.24}
\newcommand{\pairedLo}{0.15}
\newcommand{\pairedHi}{0.34}
\newcommand{\aucKtwo}{0.66}
\newcommand{\aucKsix}{0.73}
\newcommand{\eceGap}{0.080}
\newcommand{\maxSavingPct}{17}
\newcommand{\savingRthreePct}{13}
\newcommand{\coverageRthreePct}{35}
\newcommand{\nTopics}{5}
\newcommand{\nTopicsGapWins}{5}
\newcommand{\topicGapAucMin}{0.68}
\newcommand{\topicGapAucMax}{0.87}
\newcommand{\propLevelAuc}{0.71}
\newcommand{\propLevelAucHuman}{0.63}
\newcommand{\propContrastAucHuman}{0.78}
\newcommand{\propLevelAucLo}{0.67}
\newcommand{\propLevelAucHi}{0.75}
\newcommand{\propContrastAuc}{0.87}
\newcommand{\propContrastAucLo}{0.85}
\newcommand{\propContrastAucHi}{0.90}
\newcommand{\propDeltaAuc}{0.16}
\newcommand{\propDeltaAucLo}{0.13}
\newcommand{\propDeltaAucHi}{0.19}
\newcommand{\propContrastAucNonUS}{0.87}
\newcommand{\propContrastAucTestPairs}{0.88}
\newcommand{\propLevelAucTestPairs}{0.72}
\newcommand{\propDeltaAucTestPairs}{0.16}
\newcommand{\nTestPairs}{1151}
\newcommand{\nTestCountries}{5}

\newcommand{\propAucEpsTight}{0.90}
\newcommand{\propAucEpsLoose}{0.81}

\newcommand{\propContrastQHeldout}{0.86}
\newcommand{\propLevelQHeldout}{0.73}
\newcommand{\propDeltaQHeldout}{0.14}
\newcommand{\propLevelErr}{0.199}
\newcommand{\propContrastErr}{0.166}
\newcommand{\propInvalidFullPct}{63.7}
\newcommand{\propInvalidEightyPct}{54.8}
\newcommand{\propRiskRedPct}{8.9}
\newcommand{\propRiskRedLoPct}{7.5}
\newcommand{\propRiskRedHiPct}{10.3}
\newcommand{\propAnsSupPct}{24}
\newcommand{\propAnsConPct}{25}
\newcommand{\propAnsUnrPct}{51}
\newcommand{\propRiskRedTestPct}{8.8}
\newcommand{\nContrast}{14388}
\newcommand{\nNonUS}{11990}
\newcommand{\nQTestProp}{3094}
\newcommand{\nQTestPropQ}{24}
\newcommand{\propNSupported}{2816}
\newcommand{\propNContradicted}{4658}
\newcommand{\propNUnresolved}{6914}
\newcommand{\propRawResolvedInvalidPct}{62.3}
\newcommand{\propSupportedPct}{20}
\newcommand{\propContradictedPct}{32}
\newcommand{\propUnresolvedPct}{48}
\newcommand{\focalLevelAuc}{0.35}
\newcommand{\focalContrastAuc}{0.61}
\newcommand{\sharedErrPct}{58}

\newcommand{\sharedVaryPct}{21}

\newcommand{\constOfSharedPct}{65}
\newcommand{\sharedErrLoPct}{53}
\newcommand{\sharedErrHiPct}{65}
\newcommand{\constOfSharedLoPct}{56}
\newcommand{\constOfSharedHiPct}{72}
\newcommand{\cfNCal}{7}
\newcommand{\cfNTest}{5}

\newcommand{\cfMaeBefore}{0.194}
\newcommand{\cfMaeAfter}{0.129}
\newcommand{\cfMaeAfterPerm}{0.246}

\newcommand{\cfDmaeLo}{0.044}
\newcommand{\cfDmaeHi}{0.087}
\newcommand{\cfAucBefore}{0.72}
\newcommand{\cfAucAfter}{0.71}

\newcommand{\cfDaucLo}{-0.05}
\newcommand{\cfDaucHi}{0.05}
\newcommand{\cfBinLoCanc}{0.26}
\newcommand{\cfBinLoRepair}{0.019}
\newcommand{\cfBinMidCanc}{0.62}
\newcommand{\cfBinMidRepair}{0.052}
\newcommand{\cfBinHiCanc}{0.86}
\newcommand{\cfBinHiRepair}{0.123}
\newcommand{\sigD}{0.87}
\newcommand{\sigW}{0.88}
\newcommand{\sigEntropy}{0.65}
\newcommand{\sigMargin}{0.63}
\newcommand{\sigLomo}{0.86}

\newcommand{\contrastAucFamRep}{0.88}

\newcommand{\sigDvsEntropy}{0.23}
\newcommand{\gasLevelAuc}{0.71}
\newcommand{\gasContrastAuc}{0.87}
\newcommand{\gasDauc}{0.16}
\newcommand{\gasConstPct}{58}
\newcommand{\gasDaucLo}{0.13}
\newcommand{\gasDaucHi}{0.20}

\newcommand{\wvsLevelAuc}{0.73}
\newcommand{\wvsContrastAuc}{0.87}
\newcommand{\wvsDauc}{0.14}
\newcommand{\wvsConstPct}{88}
\newcommand{\wvsDaucLo}{0.02}
\newcommand{\wvsDaucHi}{0.23}

}{}
\providecommand{\nQuestions}{60}\providecommand{\nCountries}{12}
\providecommand{\nSimulators}{4}
\providecommand{\aucD}{--}

\providecommand{\invalidFullPct}{--}\providecommand{\invalidEightyPct}{--}
\providecommand{\pInvLowD}{--}\providecommand{\pInvHighD}{--}

\providecommand{\rateDirection}{--}\providecommand{\rateMean}{--}\providecommand{\rateGap}{--}\providecommand{\rateTV}{--}
\providecommand{\nCal}{--}\providecommand{\nTest}{--}
\providecommand{\sharedErrPct}{58}\providecommand{\constOfSharedPct}{64}
\providecommand{\sharedVaryPct}{21}

\providecommand{\propLevelQHeldout}{0.64}\providecommand{\propDeltaQHeldout}{0.14}
\providecommand{\propRawResolvedInvalidPct}{56.4}
\providecommand{\propContrastAucTestPairs}{0.78}\providecommand{\nTestPairs}{1614}\providecommand{\nTestCountries}{5}
\providecommand{\propAucEpsTight}{0.83}\providecommand{\propAucEpsLoose}{0.78}

\providecommand{\sharedErrLoPct}{53}\providecommand{\sharedErrHiPct}{65}\providecommand{\constOfSharedLoPct}{56}\providecommand{\constOfSharedHiPct}{72}
\providecommand{\propRiskRedPct}{8.9}\providecommand{\propRiskRedLoPct}{7.5}\providecommand{\propRiskRedHiPct}{10.3}\providecommand{\propRiskRedTestPct}{8.8}
\providecommand{\propAnsSupPct}{24}\providecommand{\propAnsConPct}{25}\providecommand{\propAnsUnrPct}{51}

\providecommand{\gasDaucLo}{0.13}\providecommand{\gasDaucHi}{0.20}\providecommand{\wvsDaucLo}{0.02}\providecommand{\wvsDaucHi}{0.23}
\providecommand{\cfNCal}{7}\providecommand{\cfNTest}{5}
\providecommand{\cfMaeBefore}{0.194}\providecommand{\cfMaeAfter}{0.129}
\providecommand{\cfDmaeLo}{0.044}\providecommand{\cfDmaeHi}{0.087}
\providecommand{\cfAucBefore}{0.63}\providecommand{\cfAucAfter}{0.67}
\providecommand{\cfDaucLo}{-0.02}\providecommand{\cfDaucHi}{0.09}
\providecommand{\cfBinLoCanc}{0.26}\providecommand{\cfBinLoRepair}{0.019}
\providecommand{\cfBinMidCanc}{0.62}\providecommand{\cfBinMidRepair}{0.052}
\providecommand{\cfBinHiCanc}{0.86}\providecommand{\cfBinHiRepair}{0.123}
\providecommand{\sigD}{0.78}\providecommand{\sigW}{0.73}\providecommand{\sigEntropy}{0.60}
\providecommand{\sigMargin}{0.60}\providecommand{\sigLomo}{0.76}
\providecommand{\contrastAucFamRep}{0.77}
\providecommand{\gasLevelAuc}{0.62}\providecommand{\gasContrastAuc}{0.77}\providecommand{\gasDauc}{0.15}\providecommand{\gasConstPct}{58}
\providecommand{\wvsLevelAuc}{0.67}\providecommand{\wvsContrastAuc}{0.82}\providecommand{\wvsDauc}{0.15}\providecommand{\wvsConstPct}{88}
\providecommand{\cfMaeAfterPerm}{0.269}

\providecommand{\sigDvsEntropy}{0.18}
\providecommand{\propLevelAucTestPairs}{0.63}\providecommand{\propDeltaAucTestPairs}{0.15}
\providecommand{\propLevelAucHuman}{0.63}\providecommand{\propContrastAucHuman}{0.78}

\title{Wrong Together: Shared Bias in User Simulation}

\author{%
  Saber Zerhoudi\\
  University of Passau\\
  Passau, Germany\\
  \texttt{szerhoudi@acm.org}
}

\begin{document}
\sloppy   
\maketitle

\begin{abstract}
User simulators generate responses in place of human participants, allowing researchers to estimate the result of a study. To use these estimates before collecting human responses, researchers need a way to assess their reliability. We test whether disagreement among simulators can identify unreliable estimates of a group's response proportions and of differences between groups. We evaluate this signal using \nSimulators{} language-model simulators and existing GlobalOpinionQA responses for \nQuestions{} questions in \nCountries{} countries. Disagreement predicts invalidity more accurately for differences between two groups than for a single group's response proportions (AUC $\propContrastAuc$ versus $\propLevelAuc$). Errors shared by the simulators remain invisible to this signal. When such an error affects both groups equally, it cancels in the comparison, helping explain the stronger result for group differences. We test this explanation by estimating the shared bias using one set of groups and subtracting it from predictions for a separate set. This correction reduces prediction error without changing disagreement among the simulators. In a retrospective analysis, we simulate referring the $20\%$ of group comparisons with the greatest disagreement for human assessment. Among the remaining comparisons with a conclusive validity judgment, $\propInvalidEightyPct\%$ are judged invalid, compared with $\propInvalidFullPct\%$ before referral.\footnote{Code and simulator outputs: \url{https://github.com/searchsim-org/neurips26-wrong-together}}
\end{abstract}

\section{Introduction}
Researchers use language models to generate responses on behalf of study participants and estimate how a population would respond. This \emph{silicon-sampling} approach~\citep{argyle2023out,santurkar2023whose} can reduce the work of collecting human responses, provided the simulated estimates support the conclusions researchers would draw from human data. We focus on survey respondents, complementing work on simulated users of search, recommender, and dialogue systems~\citep{sim2real2026,walkaway2026,zerhoudi2026tour}. A survey study may estimate the proportion choosing an option in one population, which we call a \emph{level}, or the difference in that proportion between \mbox{two populations, a \emph{contrast}}.

Existing evaluations show why these estimates need checking. Simulators for agentic tasks are more cooperative and more uniform in style than human participants, overstating measured agent performance~\citep{sim2real2026}. Similar differences appear in purchase decisions, where simulated shoppers reproduce buyers' decisions more closely than non-buyers' decisions~\citep{walkaway2026}. HumanStudy-Bench assesses simulator fidelity at the study level by comparing simulated estimates and effect sizes with reconstructed human studies while accounting for sampling uncertainty~\citep{humanstudybench2026}. Each comparison uses an available human reference to measure simulator error.

For a new study, researchers may need to decide which simulated estimates require human assessment before that reference is available. Disagreement among several simulators could identify unreliable estimates, although agreement can conceal an error shared by all models. \emph{Can simulator disagreement predict invalid estimates, and does its usefulness differ between levels and contrasts?}

We examine this question using \nQuestions{} GlobalOpinionQA questions~\citep{durmus2023global}, \nCountries{} countries, and \nSimulators{} simulators. Disagreement predicts invalidity more accurately for contrasts than for levels. The error decomposition identifies a component shared across simulators and constant across countries, which cancels when two countries are compared.

We define invalidity using an equivalence margin and uncertainty from both human samples and simulator draws (Section~\ref{sec:formulation}). We test the shared-error explanation by estimating a question-specific bias on calibration countries and correcting level estimates on separate test countries (Section~\ref{sec:mech}). This correction uses human calibration data, while the disagreement score itself uses simulator outputs alone. We then evaluate whether referring high-disagreement contrasts reduces invalidity among the estimates retained for simulation (Section~\ref{sec:referral}). All evaluations use existing survey responses.

\section{Related Work}
User simulators support the training and evaluation of dialogue systems and recommenders, using statistical user models~\citep{schatzmann2006survey}, configurable probabilistic simulators~\citep{ie2019recsim}, and prompted language models~\citep{balog2025theory}. Evaluating these simulators requires comparing their behavior with that of human users. Such comparisons find greater cooperation and less stylistic variation among simulators~\citep{sim2real2026}, a disengagement deficit across model families~\citep{walkaway2026}, and a need to align simulated responses with retail users~\citep{pruksachatkun2026salessim}. In survey simulation, the corresponding comparison concerns response distributions. Persona-conditioned LLMs approximate human response distributions with mixed success~\citep{argyle2023out,santurkar2023whose,bernard2026llm}, representing some populations more closely than others~\citep{santurkar2023whose,durmus2023global}. GlobalOpinionQA provides the response distributions we use to examine whether several simulators share these errors across countries.

For studies using simulated participants, the relevant question extends beyond the realism of individual responses to the accuracy of the quantities estimated from them. HumanStudy-Bench evaluates estimands and effect sizes against reconstructed human studies, accounting for human sampling uncertainty~\citep{humanstudybench2026}. Response realism and treatment-effect accuracy can also diverge~\citep{liji2026treatment}. The distinction between response realism and effect accuracy motivates evaluating levels and contrasts separately. We examine whether simulator outputs identify invalid estimates for each, without using the target populations' human responses.

Predicting invalidity without the target human responses requires a signal from the simulators themselves. Ensemble disagreement is an established uncertainty signal~\citep{lakshminarayanan2017simple}, and \citet{gorbett2026disagreement} use cross-model disagreement to detect errors without labels in question answering. Using such a signal requires separating agreement from validity, a distinction also made in work on LLM judges~\citep{chehbouni2025neither}. We study the corresponding problem for simulated population estimates: models may agree because they share an error, while a between-group difference may cancel part of that error. We use an equivalence test to define acceptable error~\citep{lakens2017equivalence} and selective prediction to refer high-disagreement estimates for \mbox{human assessment~\citep{elyaniv2010foundations}}.

The referral rule and any correction of simulated estimates must themselves be checked against human evidence. \citet{hullman2026validating} argue that simulator corrections need these checks to support confirmatory research, while \citet{persson2026statistical} caution that validation on past studies cannot certify performance on new ones. Our tests therefore separate countries and questions within existing surveys and limit the resulting conclusions to that setting.

\section{Problem and Method}
\label{sec:formulation}
\label{sec:method}
Our primary estimates concern response proportions. A level is the proportion choosing a particular option in a population. A contrast subtracts the corresponding proportions in two populations. Let $j\in\mathcal{J}$ index one such quantity, or estimand, with human population value $\theta_{H,j}$ and simulator value $\theta_{s,j}$. We report the mean of $K$ simulators, $\bar\theta_j=\frac1K\sum_s\theta_{s,j}$, and assess its error:
\begin{equation}
\delta_{j} = \bar\theta_{j} - \theta_{H,j}.
\end{equation}
We observe finite-sample estimates $\hat\theta_{H,j}$ and $\hat\theta_{s,j}$. For each estimand, the validity decision concerns the ensemble estimate. We ask whether variation among the simulator estimates predicts that decision without using the target human responses $D_{H,j}$.

Determining whether an estimate meets an error margin requires accounting for sampling uncertainty. For example, an error of $0.07$ is more precisely determined when the human standard error is $0.01$ than when it is $0.12$. We define validity by $\delta_j\in[-\epsilon_j,\epsilon_j]$, where $\epsilon_j$ is a scientifically meaningful equivalence margin. We fix the margin independently of the confidence interval. Setting it from the interval width would make a less precise human reference easier to match.

For an estimated error $\hat\delta$ with standard error $\mathrm{SE}$, we use a conservative $95\%$ confidence-interval rule ($z=1.96$). Comparing the bounds $|\hat\delta|\pm z\,\mathrm{SE}$ with the margin gives three decisions:
\begin{itemize}\itemsep2pt
\item \textbf{supported}: the interval lies inside the equivalence region
      ($|\hat\delta|+z\,\mathrm{SE}\le\epsilon_j$);
\item \textbf{contradicted}: it lies outside ($|\hat\delta|-z\,\mathrm{SE}>\epsilon_j$);
\item \textbf{unresolved}: the interval straddles $\epsilon_j$.
\end{itemize}
The standard error in these decisions combines the human sampling error and the simulator Monte-Carlo error from limited draws, $\mathrm{SE}^2=\mathrm{SE}_{\mathrm{human}}^2+\mathrm{SE}_{\mathrm{sim}}^2$, treated as independent. For a proportion, $\mathrm{SE}_{\mathrm{human}}^2=\hat p(1-\hat p)/n_H$ and $\mathrm{SE}_{\mathrm{sim}}^2=K^{-2}\sum_s \hat p_s(1-\hat p_s)/n_s$ over the $K$ simulators with $n_s$ draws each. A contrast adds the two populations' variances. The unresolved category retains cases where these samples do not establish whether the margin is met.

To predict these validity decisions, we measure cross-simulator disagreement $D_j$ as the mean absolute deviation of the $K$ simulator estimates. Within-simulator uncertainty is $W_{s,j}=\mathrm{Var}_r(\hat\theta^{(r)}_{s,j})$ over repeated stochastic simulation $r$. These measures use simulator outputs only. For the secondary estimands defined in Section~\ref{sec:setup}, we use the fraction disagreeing with the plurality vote for direction estimates and mean pairwise total-variation distance for full distributions. Continuous estimands use mean absolute deviation throughout.

We use disagreement to select estimates for human assessment. For a threshold $\tau$, we retain the simulated estimate when $D_j\le\tau$ and otherwise refer it for human assessment. Coverage is the fraction of estimates retained. Risk is the invalidity rate among resolved, retained estimates, with unresolved estimates reported separately. We evaluate both the ranking and the referral rule retrospectively against existing human responses. Ranking performance is measured by AUC, the frequency with which an invalid estimate receives greater disagreement than a valid estimate among resolved cases.

This referral rule uses disagreement directly, without fitting a predictor to human responses. To test the explanation for the level--contrast difference, a separate analysis estimates shared error using human responses from calibration countries and applies the correction to test countries. Appendix~\ref{app:decision} uses the same separation of countries to fit and evaluate probabilities of invalidity for \mbox{a calibrated referral rule}.

\section{Experimental Setup}
\label{sec:setup}
We use Global\allowbreak Opinion\allowbreak QA~\citep{durmus2023global}, which records human response distributions from the World Values Survey and Pew Global Attitudes surveys. We sample \nQuestions{} questions covering \nCountries{} of the best-covered countries: Russia, Mexico, Germany, Poland, Turkey, Pakistan, Jordan, Kenya, France, Lebanon, the United States, and Japan. Each question--country response distribution supplies one primary level estimand per eligible option. Contrasts compare the same option between two countries. Table~\ref{tab:dataset} reports the number of estimates in each analysis.

These surveys report weighted proportions without per-question respondent counts. We approximate the human standard error using each country's survey-design sample size and a multinomial distribution. Uncertainty from limited simulator draws is also material: median $\mathrm{SE}_{\mathrm{sim}}\!\approx\!0.025$, compared with $\mathrm{SE}_{\mathrm{human}}\!\approx\!0.016$. Including the simulator term moves about a fifth of the contrast estimates that would otherwise be resolved to \emph{unresolved}.

We run \nSimulators{} simulators from four model families: \texttt{gemma4-31b}, \texttt{mistral-small}, \texttt{gpt-4o-mini}, \texttt{qwen3-next-80b}, \texttt{gpt-4o}, and \texttt{mistral-large}. Each is prompted to respond as a randomly sampled person from the target country. Repeated stochastic draws provide its response distribution and within-model spread $W$. We also evaluate one model per family to assess the effect of including multiple variants from a family.

\begin{table}[!htbp]
\centering
\caption{Study scale and numbers of evaluated estimates. Primary estimates use response proportions. Secondary analyses use questions with ordered response options.}
\label{tab:dataset}
\begin{tabular*}{\linewidth}{@{\extracolsep{\fill}}llr@{}}
\toprule
Analysis & Quantity & Count \\
\midrule
Study & Questions & 60 \\
 & Populations & 12 \\
 & Simulators & 6 \\
 & Model families & 4 \\
\addlinespace
Primary: proportions & Option-level estimates & 2616 \\
 & Pairwise contrasts & 14388 \\
\addlinespace
Secondary: ordinal & Direction & 720 \\
 & Mean & 720 \\
 & Ordinal contrasts & 660 \\
 & Full distributions & 720 \\
\bottomrule
\end{tabular*}

\end{table}

For these data, each primary evaluation concerns one ensemble estimate per question, country, and option for levels, or per question, country pair, and option for contrasts. We weight each option by $1/m_q$, where $m_q$ is the question's number of eligible options, so each question has equal total weight. Inference is clustered by question because its response proportions and country contrasts are dependent. Table~\ref{tab:difficulty} separately evaluates \mbox{individual simulator--estimand pairs}.

We apply an equivalence margin of $0.10$ to these primary proportions and contrasts, corresponding to a ten-percentage-point error. Margins are fixed before the human comparison and independently of sampling uncertainty. We report invalidity among resolved estimates and the \mbox{unresolved share separately}.

We extend the comparison to ordered response options through four secondary estimands: \emph{direction} (the modal non-refusal option), \emph{mean} (the mean response index), \emph{ordinal contrast} (the difference in mean index from the United States), and \emph{distribution} (total-variation distance between response distributions). For these secondary analyses, we use margins of $0.30$ response-index units for means and ordinal contrasts, $0.15$ total-variation distance for distributions, and exact modal agreement for direction. The index scales span about four units. Appendix~\ref{sec:ordinal} reports these \mbox{analyses and their sensitivity checks}.

We evaluate the primary signal over all pairwise contrasts, on \nQTestPropQ{} separate test questions, and on contrasts restricted to test countries. For analyses requiring human calibration, we split the \nCountries{} countries into \nCal{} calibration countries $\mathcal{C}_{\mathrm{cal}}$ and \nTest{} test countries $\mathcal{C}_{\mathrm{test}}$. Correction offsets and learned probability mappings use only calibration-country human responses. Prompts, temperatures, model identifiers, draw counts, and parsing procedures will be released with the code.

\section{Results}
\label{sec:results}
\subsection{Error prediction for levels and contrasts}
Disagreement distinguishes invalid from valid ensemble estimates more accurately for contrasts than for levels. AUC is $\propContrastAuc$ ($95\%$ CI $[\propContrastAucLo,\propContrastAucHi]$) for contrasts and $\propLevelAuc$ ($[\propLevelAucLo,\propLevelAucHi]$) for levels. The paired difference is $\propDeltaAuc$ (question-clustered CI $[\propDeltaAucLo,\propDeltaAucHi]$). The contrast result also appears on $\nQTestPropQ$ separate test questions. Across $\nQTestProp$ corresponding contrast estimands, AUC is $\propContrastQHeldout$. Excluding the United States from both sides of a contrast gives \mbox{AUC $\propContrastAucNonUS$ across $\nNonUS$ contrasts}.

\begin{figure}[t]
\centering
\includegraphics[width=.92\textwidth]{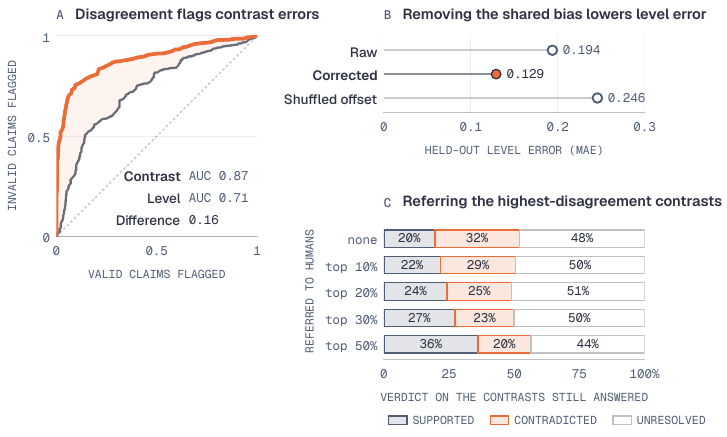}
\caption{\textbf{Error prediction, bias correction, and selective referral.} (A) Disagreement predicts error better for
contrasts than absolute levels ($\Delta\mathrm{AUC}=\propDeltaAuc$,
question-clustered $95\%$ CI). (B) Each point is test level error (MAE) on the test countries: before correction
($\cfMaeBefore$), after subtracting a per-question bias estimated on the $\cfNCal$ calibration
countries ($\cfMaeAfter$), and after subtracting an offset from unrelated questions as a placebo
($\cfMaeAfterPerm$). Only the real correction lowers the error.
(C) Verdicts of the equivalence test on the contrasts that simulation still answers, as the
highest-disagreement share is referred to humans. Referral shrinks the contradicted share, but
about half of the answered contrasts stay unresolved at every threshold, so referral orders the
human checks without certifying the rest. Among resolved contrasts, invalidity falls from
$\propInvalidFullPct\%$ when simulation answers all of them to $\propInvalidEightyPct\%$ when it
answers the lowest-disagreement $80\%$.}
\label{fig:primary}
\end{figure}

\begin{table}[t]
\centering
\caption{\textbf{Level and contrast AUC on response-option proportions.} Prediction is evaluated on all pairs, test questions, and test-population pairs. Referral retains the lowest-disagreement $80\%$ of contrasts. AUC is question-weighted, with question-clustered $95\%$ confidence intervals.
Validity uses a fixed equivalence margin.}
\label{tab:proportion}
\begin{threeparttable}
\small
\begin{tabular}{@{}l r r r@{}}
\toprule
Evaluation & Absolute level & Contrast & Difference\tnote{d} \\
\midrule
All pairwise contrasts & \propLevelAuc\ [\propLevelAucLo, \propLevelAucHi]
  & \propContrastAuc\ [\propContrastAucLo, \propContrastAucHi]
  & $+$\propDeltaAuc\ [\propDeltaAucLo, \propDeltaAucHi] \\
Test-question AUC & \propLevelQHeldout & \propContrastQHeldout & $+$\propDeltaQHeldout \\
Test-population pairs only & \propLevelAucTestPairs & \propContrastAucTestPairs
  & $+$\propDeltaAucTestPairs\ \tnote{a} \\
Reference rotated (non-US) & --- & \propContrastAucNonUS & \tnote{b} \\
Invalidity, full coverage & --- & \propInvalidFullPct\% & \tnote{c} \\
Invalidity, $80\%$ coverage & --- & \propInvalidEightyPct\% & \tnote{c} \\
\bottomrule
\end{tabular}
\begin{tablenotes}[flushleft]\footnotesize
\item[a] Over the \nTestPairs\ contrasts among the \nTestCountries\ test populations. No predictor is refit.
\item[b] Over the \nNonUS\ contrasts that never involve the United States, reference
  rotated across all pairs.
\item[c] Question-weighted over all \nContrast\ contrasts. Raw decisions:
  \propNSupported\ supported, \propNContradicted\ contradicted, \propNUnresolved\
  unresolved. Question-weighting makes \propInvalidFullPct\% differ from the raw ratio
  \propRawResolvedInvalidPct\%.
\item[d] Inference resamples questions, not contrast rows. The primary $\Delta$AUC
  interval $[\propDeltaAucLo,\propDeltaAucHi]$ and both per-source intervals exclude zero.
\end{tablenotes}
\end{threeparttable}

\end{table}

These test questions retain a level--contrast AUC difference of $\propDeltaQHeldout$ (Table~\ref{tab:proportion}). Restricting the country pairs to the $\nTestCountries$ test populations gives level AUC $\propLevelAucTestPairs$ and contrast AUC $\propContrastAucTestPairs$ ($\Delta\mathrm{AUC}=\propDeltaAucTestPairs$). Both checks use the same disagreement score without refitting.

\paragraph{Disagreement and other uncertainty signals}
To assess whether disagreement captures more than response ambiguity, we compare it with other uncertainty measures. Cross-model disagreement (AUC $\sigD$), pooled within-model sampling spread ($\sigW$), and leave-one-model-out instability ($\sigLomo$) have similar predictive performance, with paired AUC differences within $\pm 0.02$. Each exceeds the entropy of the ensemble response distribution ($\sigEntropy$) and its plurality margin ($\sigMargin$). Disagreement exceeds \mbox{entropy by $\sigDvsEntropy$ in AUC}.

The comparable performance of sampling spread depends on pooling it over the ensemble. For five of the six individual simulators, sampling spread predicts invalidity of that simulator's contrast estimates with AUC $0.51$ to $0.58$. Disagreement gives $0.75$ to $0.80$ on the same estimates. The cross-model result also persists when the ensemble contains one model per pretraining family, giving contrast AUC $\contrastAucFamRep$, close to \mbox{the full ensemble's $\propContrastAuc$}.

\subsection{Shared error and bias correction}
\label{sec:mech}
To examine why contrast invalidity is more predictable, we separate shared error from simulator-specific deviations. Write $\theta_{s,j}=\theta_{H,j}+b_j+\iota_{s,j}$ with $\frac1K\sum_s\iota_{s,j}=0$. Here $b_j$ is the shared error and $\iota_{s,j}$ is the simulator-specific deviation. Averaging over simulators leaves ensemble error $|b_j|$. Disagreement is unchanged by the shared offset because mean absolute deviation $D_j=\mathrm{MAD}_s(\theta_{s,j})$ depends on $\{\iota_{s,j}\}_s$, while observed disagreement also includes sampling noise $W$.

Identifying the component removed by a country contrast requires a further split across populations. For question $q$, country $c$, and option $o$, write signed simulator error as $e_{s,q,c,o}=\bar a_{q,o}+p_{q,c,o}+m_{s,q,c,o}$. The component $\bar a_{q,o}$ is shared across simulators and constant across countries. The component $p_{q,c,o}$ is shared across simulators but varies across countries, and $m_{s,q,c,o}$ is the simulator-specific residual. A contrast cancels $\bar a_{q,o}$ and retains differences in $p$ and $m$.

\begin{figure}[t]
\centering
\includegraphics[width=0.65\linewidth]{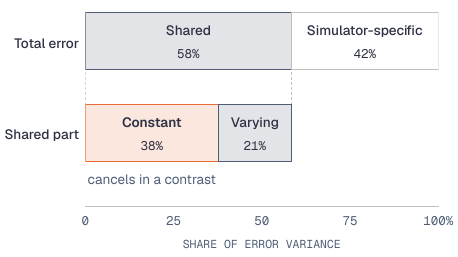}
\caption{\textbf{Shared and simulator-specific error.} $\sharedErrPct\%$ of error variance is shared across simulators, and $\constOfSharedPct\%$ of that component is constant across countries. A contrast cancels the country-constant component. Country-varying shared error and simulator-specific error remain.}
\label{fig:decomp}
\end{figure}

We estimate the components by empirical means over the crossed $(\text{simulator}\times\text{country})$ panel, weight options by $1/m_q$, and resample questions for intervals. Of total error variance, $\sharedErrPct\%$ ($95\%$ CI $[\sharedErrLoPct,\sharedErrHiPct]$) is shared across simulators. Of that shared component, $\constOfSharedPct\%$ ($[\constOfSharedLoPct,\constOfSharedHiPct]$) is constant across countries and cancels in a contrast (Figure~\ref{fig:decomp}). Cancellation leaves the country-varying shared component, which accounts for $\sharedVaryPct\%$ of original error variance. The remaining errors give mean absolute error $\propLevelErr$ for levels and $\propContrastErr$ for contrasts. Cancellation reduces error while retaining a shared component \mbox{that disagreement cannot measure}.

\paragraph{Bias correction across countries}
We test whether the country-constant component can be estimated from other countries by calculating a question- and option-specific offset on the $\cfNCal$ calibration countries:
$\bar a_{q,o}=\frac{1}{|\mathcal{C}_{\mathrm{cal}}|}\sum_{c\in\mathcal{C}_{\mathrm{cal}}} (\bar\theta_{q,c,o}-\theta_{H,q,c,o})$.
We subtract it from level estimates for the $\cfNTest$ test countries. Mean absolute error falls from $\cfMaeBefore$ to $\cfMaeAfter$ (question-clustered $95\%$ CI on the reduction $[\cfDmaeLo,\cfDmaeHi]$; Figure~\ref{fig:primary}B). Shuffling offsets across questions with the same option count raises median error to $\cfMaeAfterPerm$. The observed reduction exceeds that of every shuffled correction.

Applying the same offset to every simulator leaves disagreement scores and their ordering unchanged. AUC against the corrected validity labels changes from $\cfAucBefore$ to $\cfAucAfter$ (difference CI $[\cfDaucLo,\cfDaucHi]$). This correction reduces level error without a clear change in AUC, consistent with removing an error component that disagreement does not measure. For contrasts, subtracting a country-constant offset changes nothing because it already cancels in the difference.

\paragraph{Correction by question}
The decomposition also predicts which questions should benefit most from this correction. We group questions into thirds by the share of calibration error that is constant across countries. These shares are $\cfBinLoCanc$, $\cfBinMidCanc$, and $\cfBinHiCanc$ in the low, middle, and high groups. The corresponding reductions in test-country level error are $\cfBinLoRepair$, $\cfBinMidRepair$, and $\cfBinHiRepair$. Questions with more country-constant error on calibration countries therefore benefit more from \mbox{correction on test countries}.

\subsection{Selective referral}
\label{sec:referral}
We use the disagreement score to rank contrasts for human assessment. At full coverage, invalidity among resolved contrasts is $\propInvalidFullPct\%$. Referring the $20\%$ with the greatest disagreement and retaining the rest lowers this rate to $\propInvalidEightyPct\%$, a reduction of $\propRiskRedPct$ points (question-clustered $95\%$ CI $[\propRiskRedLoPct, \propRiskRedHiPct]$). Restricting the comparison to pairs of test countries gives a reduction of $\propRiskRedTestPct$ points.

This reduction concerns resolved estimates. Across all $\nContrast$ contrasts, the equivalence test supports $\propSupportedPct\%$, contradicts $\propContradictedPct\%$, and leaves $\propUnresolvedPct\%$ unresolved. Among the retained $80\%$, $\propAnsUnrPct\%$ remain unresolved. These decision shares are unweighted, whereas invalidity rates are question-weighted and computed \mbox{among resolved estimates}.

Disagreement therefore determines the order of human assessment. An estimate retained by the referral rule still needs an equivalence decision, and many retained estimates remain unresolved.

\subsection{Sensitivity and additional comparisons}
\paragraph{Sampling uncertainty and equivalence margins} The validity labels themselves depend on sampling uncertainty: larger within-model spread $W$ widens the interval and can change which estimates are resolved. A signal correlated with $W$ could therefore appear predictive partly because uncertainty enters the decision rule. Using human sampling uncertainty alone preserves the ordering (level AUC $\propLevelAucHuman$, contrast AUC $\propContrastAucHuman$). Scaling every margin from $0.5\times$ to $2\times$ also preserves it, with contrast AUC in $[\propAucEpsLoose,\propAucEpsTight]$ (Section~\ref{sec:robust}).

\paragraph{Results by survey source} The level--contrast ordering also persists when we analyze Pew Global Attitudes and the World Values Survey separately. Contrast and level AUC are $\gasContrastAuc$ and $\gasLevelAuc$ on Pew, and $\wvsContrastAuc$ and $\wvsLevelAuc$ on World Values. The paired differences are $\gasDauc$ ($95\%$ CI $[\gasDaucLo,\gasDaucHi]$) and $\wvsDauc$ ($[\wvsDaucLo,\wvsDaucHi]$). The country-constant component accounts for $\gasConstPct\%$ and $\wvsConstPct\%$ of shared error, respectively. The World Values interval is wider because it covers fewer questions.

\paragraph{One option per question}
Our primary analysis includes all eligible options with equal total weight per question. To check sensitivity to the set of options, we select the option the simulator ensemble chooses most often, without using human responses, and retain only that option for each question. Contrast AUC is $\focalContrastAuc$, compared with $\focalLevelAuc$ for levels. The ordering is preserved, although the below-chance level AUC may reflect both shared bias and selection of a modal option from simulator outputs. This dependence on option selection limits what the check establishes about the mechanism.

\paragraph{Ordinal estimands}
Extending the analysis from response proportions to ordered response indices gives contrast AUC $\aucGap$ ($95\%$ CI $[\aucGapLo,\aucGapHi]$), compared with $\aucMean$ for absolute means and $\aucTV$ for full distributions. The ordering persists across equivalence margins, assumed human sample sizes, survey sources, ensemble sizes, and population splits (Appendix~\ref{sec:ordinal}).

\section{Conclusion}
Across \nQuestions{} survey questions, \nCountries{} countries, and \nSimulators{} simulators, disagreement predicts invalidity more accurately for between-group differences (AUC $\propContrastAuc$) than for a single group's response proportions ($\propLevelAuc$). The decomposition and correction on separate countries connect this difference to shared error: contrasts cancel a country-constant component that disagreement cannot measure. Using disagreement to refer the highest-scoring $20\%$ of contrasts reduces invalidity among resolved, retained contrasts from $\propInvalidFullPct\%$ to $\propInvalidEightyPct\%$. This use of disagreement therefore depends on the quantity a study estimates and prioritizes the human checks that remain necessary.

\section{Limitations}
\label{sec:limitations}
Separating calibration and test populations prevents the fitted corrections and probability mappings from using test-country responses. However, the survey responses predate our study design, and the evaluation does not constitute a preregistered prediction of newly collected outcomes. The $\nSplits$ population splits assess sensitivity to the chosen countries, rather than transfer to a new human study. A prospective evaluation would fix simulator predictions and the analysis before collecting human responses. We plan to release a time-stamped protocol and a blind collection application \mbox{for such a test}.

Referral reduces risk without establishing validity for every retained estimate. In Figure~\ref{fig:primary}C, the contradicted share falls from $32\%$ at full coverage to $20\%$ at half coverage, while $44\%$ to $51\%$ remain unresolved. At $80\%$ coverage, the unweighted decision shares are $\propAnsSupPct\%$ supported, $\propAnsConPct\%$ contradicted, and $\propAnsUnrPct\%$ unresolved. These shares describe all retained estimates, whereas the headline invalidity rate uses resolved estimates and question weighting.

The surveys report weighted response proportions without per-question respondent counts. We use each country's survey-design sample size and do not adjust for the design effect of weighting. These approximations affect the human standard error and the validity decisions. In the ordinal sensitivity analysis, varying assumed sample sizes sixteen-fold changes the unresolved share from $\unresolvedNlo\%$ to $\unresolvedNhi\%$. Invalidity among resolved estimates remains between $56\%$ and $58\%$, and contrast AUC remains between $0.69$ and $0.76$ (Figure~\ref{fig:robust}B,C).

The primary evaluation covers global opinion, with \nSimulators{} simulators from four model families. Transfer to interactive tasks requires separate evidence. Appendix~\ref{app:interactive} reports an exploratory persuasion-dialogue analysis whose outcomes do not permit evaluating the disagreement predictor. Using a simulator for adaptive optimization introduces the further problem of optimizing an imperfect proxy~\citep{gao2023scaling}. Our static evaluation does not assess validity under that distribution shift.

In interactive simulation, a bias that is constant across systems would cancel in their difference. Whether a simulator's bias is sufficiently similar across systems must be tested. In conversational search, a prompted simulator over-states retrieval effectiveness on the turns where real users find little~\citep{zerhoudi2026autosim}.

Our country-based prompts also raise a representational concern. Conditioning a simulator on a country elicits the model's picture of that country's population, which can reproduce stereotypes. We use these answers only to measure their error against real respondents and do not report them as \mbox{descriptions of any population}.

\bibliographystyle{plainnat}
\bibliography{references}

\appendix

\section{Ordinal-family analysis and robustness}
\label{sec:ordinal}
\begin{figure}[H]
\centering
\begin{subfigure}{0.49\textwidth}\centering
  \includegraphics[width=0.92\linewidth]{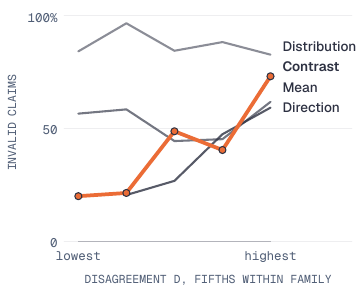}
  \caption{Invalid share by disagreement.}\label{fig:scatter}
\end{subfigure}\hfill
\begin{subfigure}{0.49\textwidth}\centering
  \includegraphics[width=0.92\linewidth]{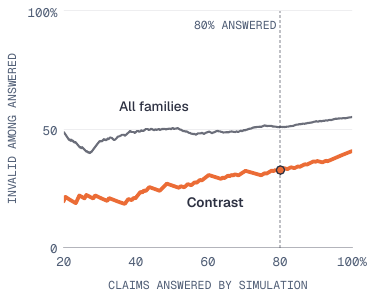}
  \caption{Risk--coverage after referral.}\label{fig:rc}
\end{subfigure}
\caption{\textbf{Selective simulation on test populations.} (a) Share of invalid estimates in each fifth of disagreement, ranked within family so the
families share one axis. The trend differs by family: contrast invalidity rises from the
lowest to the highest fifth, distribution invalidity does not. (b) Referring the
highest-disagreement estimates lowers invalidity among the answered contrast estimates; the same
rule gives a smaller reduction across all families combined. The dotted line marks $80\%$
coverage.}
\label{fig:main}
\end{figure}

\begin{figure}[H]
\centering
\begin{subfigure}{0.49\textwidth}\centering
  \includegraphics[width=0.92\linewidth]{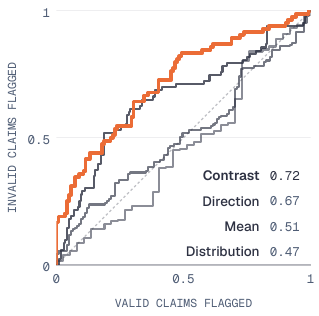}
  \caption{ROC of disagreement, by family.}\label{fig:famauc}
\end{subfigure}\hfill
\begin{subfigure}{0.49\textwidth}\centering
  \includegraphics[width=0.92\linewidth]{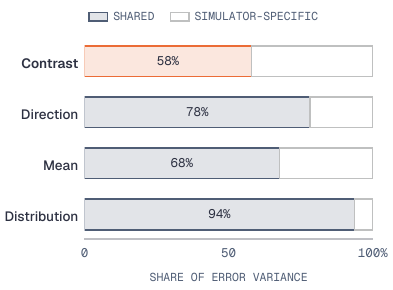}
  \caption{Shared vs.\ simulator-specific error.}\label{fig:mech}
\end{subfigure}
\caption{\textbf{Predictability and mechanism by family.} (a) ROC curves, one per estimand
family, using disagreement to catch invalid estimates: contrasts and direction rise above the
diagonal, means and distributions stay near it (AUC is shown for each family, with a dotted line at $0.5$).
(b) Fraction of estimand-error variance shared across simulators (invisible to disagreement)
versus simulator-specific. Contrast error contains a larger simulator-specific
component than mean or distribution error. Family-level invalidity rates are in Table~\ref{tab:difficulty}.}
\label{fig:second}
\end{figure}

We repeat the analysis on questions with ordered response options, using direction, mean response index, ordinal contrast, and full response distribution as estimands. Figures~\ref{fig:scatter} and~\ref{fig:famauc} show how invalidity varies with disagreement and how well disagreement ranks invalid estimates on test populations. AUC is $\aucDirection$ for direction and $\aucGap$ for ordinal contrasts ($95\%$ CI $[\aucGapLo,\aucGapHi]$, Spearman $\spearmanGap$), compared with $\aucMean$ for means and $\aucTV$ for distributions.

These differences make the estimand family important to interpreting the signal. Pooling all families gives AUC $\aucD$. Within-simulator spread $W$ gives contrast AUC $\aucWGap$ in this ordinal analysis, differing from the predictive pooled spread in the primary response-proportion analysis. We therefore report the secondary results by family.

\paragraph{Selective referral on ordinal items}
The stronger contrast prediction supports evaluating referral within that family. Invalidity falls from $\gapInvalidFullPct\%$ at full coverage to $\gapInvalidEightyPct\%$ when the $20\%$ highest-disagreement contrasts are referred (Figure~\ref{fig:rc}, red). Applying the same rule across all families gives a smaller reduction, from $\invalidFullPct\%$ to $\invalidEightyPct\%$, consistent with the weaker prediction for means and distributions. The fixed $80\%$ coverage illustrates the rule's effect. Appendix~\ref{app:decision} fits probabilities of invalidity (test-population ECE $\eceGap$) and uses assigned error and referral costs to choose a threshold, with a maximum modeled saving of $\maxSavingPct\%$.

\paragraph{Results by topic}
\label{app:domains}
To examine whether the contrast--mean ordering depends on subject matter, we assign questions to topics by keyword and evaluate the two estimands within each topic (Table~\ref{tab:domains}). Among the $\nTopics$ topics with at least $40$ contrast estimands, contrast AUC exceeds mean AUC in all $\nTopicsGapWins$, ranging from $\topicGapAucMin$ to $\topicGapAucMax$. A separate three-question group falls below the $40$-estimand threshold. The comparison tests consistency across topics within the survey domain.

\begin{table}[!htbp]
\centering
\begin{minipage}[t]{0.42\textwidth}
\centering
\caption{Selective simulation on the ordinal contrast family (robustness),
test populations. The primary proportion
\mbox{result appears in Table~\ref{tab:proportion}}.}\label{tab:selective}
\small\setlength{\tabcolsep}{4pt}
\begin{tabular}{lr}
\toprule
Ordinal contrast family & Value \\
\midrule
AUC (disagreement) & 0.72 [0.64, 0.79] \\
AUC (within-sim $W$) & 0.50 \\
AUC (all families) & 0.59 \\
invalidity, full cov.\ & 0.408 \\
invalidity, 80\% cov.\ & 0.329 \\
$P(\text{inv}\mid\text{low }D)$ & 0.456 \\
$P(\text{inv}\mid\text{high }D)$ & 0.673 \\
\bottomrule
\end{tabular}

\end{minipage}\hfill
\begin{minipage}[t]{0.56\textwidth}
\centering
\caption{Disagreement predicts contrast error, but not mean error, within
each subject-matter domain.}\label{tab:domains}
\small\begin{threeparttable}
\setlength{\tabcolsep}{3pt}
\begin{tabular}{l S[table-format=2.0] S[table-format=3.0] S[table-format=1.2] S[table-format=1.2]}
\toprule
{Topic domain} & {\#q} & {$n$} & {contrast AUC} & {mean AUC} \\
\midrule
religion/morals & 5 & 55 & 0.87 & 0.66 \\
economy/work & 12 & 132 & 0.69 & 0.56 \\
governance & 17 & 187 & 0.68 & 0.50 \\
international & 18 & 198 & 0.73 & 0.48 \\
society/tech\tnote{a} & 3 & 33 & 0.23 & 0.38 \\
other & 5 & 55 & 0.76 & 0.46 \\
\bottomrule
\end{tabular}
\begin{tablenotes}\footnotesize
\item[a] Under-powered: fewer than 40 gap estimands.
\end{tablenotes}
\end{threeparttable}

\end{minipage}
\end{table}

\paragraph{Invalidity at low disagreement}
The retained risk is also apparent among low-disagreement estimates: invalidity is $\pInvLowD$, compared with $\pInvHighD$ among high-disagreement estimates. For distribution estimates, disagreement and error are weakly negatively related. Both findings illustrate the limitation of using agreement to assess errors shared across simulators.

\paragraph{Shared error and sampling variance}
Figure~\ref{fig:mech} reports shared components of $\sharedFracGap$ of ordinal contrast-error variance and $\sharedFracTV$ of distribution-error variance. These are distinct from the primary response-proportion decomposition in Section~\ref{sec:mech}.

This shared component also limits the benefit of additional simulator draws. For a fixed estimand, mean squared error separates into squared bias and sampling variance. Additional draws reduce the sampling component $W$ but do not remove systematic bias (Table~\ref{tab:biasvar}). More draws can therefore improve the precision of a simulated estimate without bringing it closer \mbox{to the human reference}.

\paragraph{Invalidity by estimand} Table~\ref{tab:difficulty} reports invalidity for direction ($\rateDirection$), contrasts ($\rateGap$), means ($\rateMean$), and distributions ($\rateTV$). Contrasts have lower invalidity than the means from which they are constructed. Adjusting for human-reference signal-to-noise, effect size, survey source, and simulator preserves this ordering. On the standardized error scale $|\delta|/\sigma_H$, contrast error is $0.262$ lower than mean error (SE $0.043$, question-clustered $p<0.001$). These checks are consistent with shared-bias cancellation contributing to the difference.

\paragraph{Invalidity by model}
The high invalidity rates also persist across the included model groups. For \texttt{gpt-4o} and \texttt{mistral-large}, invalidity is $\invalidFrontierPct\%$, compared with $\invalidSmallPct\%$ for the smaller models. These comparisons do not isolate parameter count, model family, or training, and therefore do not establish a \mbox{causal effect of scale}.

\begin{figure}[H]
\centering
\includegraphics[width=\textwidth]{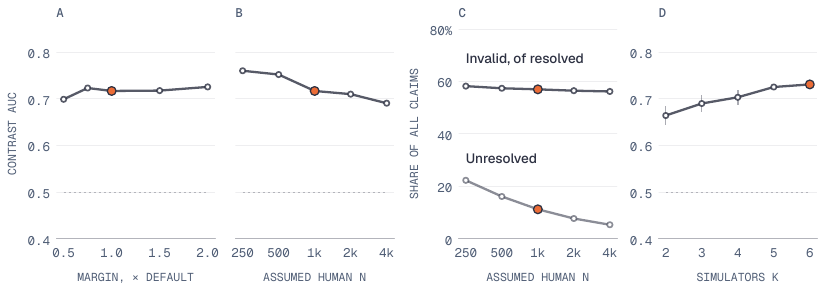}
\caption{\textbf{The contrast signal is stable and grows with the ensemble.} (A) Contrast
AUC against the equivalence margin. (B) Contrast AUC against the assumed human sample size.
(C) Verdicts on all estimates against the same sample size: the invalid share among resolved
estimates stays between $56\%$ and $58\%$, while the unresolved share falls from $\unresolvedNlo\%$
to $\unresolvedNhi\%$ as the assumed human study grows. An uncertain respondent count therefore
changes how many estimates are resolved more than how the resolved ones split. (D) Contrast AUC against the
number of simulators $K$ (bars are standard deviations over random $K$-subsets). Filled markers
are the defaults. The dashed line is $0.5$.}
\label{fig:robust}
\end{figure}

\paragraph{Sensitivity to margins and sample sizes}
\label{sec:robust}
To assess dependence on the validity criteria, we vary the ordinal-contrast equivalence margin from half to twice its default. AUC remains in $[\gapAucEpsMin,\gapAucEpsMax]$ (Figure~\ref{fig:robust}). Varying assumed human sample sizes from $250$ to $4000$ gives AUC in $[\gapAucNMin,\gapAucNMax]$, while the unresolved share falls from $\unresolvedNlo\%$ to $\unresolvedNhi\%$. Separate survey-source analyses give contrast AUC $\aucGapGAS$ on Global Attitudes and $\aucGapWVS$ on World Values. Table~\ref{tab:domains} \mbox{reports results by topic}.

\paragraph{Population splits and ensemble size}
We also repeat the calibration/test split to assess dependence on the chosen countries. Over $\nSplits$ random calibration/test splits of the populations, the test contrast AUC has median $\splitAucMed$ ($5$--$95\%$: $[\splitAucLo,\splitAucHi]$) and exceeds $0.5$ in $\splitFracAbovePct\%$ of them. Evaluating one test country at a time gives AUC between $\locoMin$ ($\locoMinCountry$) and $\locoMax$. Resampling questions also preserves the contrast--mean AUC difference, estimated at $\pairedDiff$ ($95\%$ CI $[\pairedLo,\pairedHi]$).

Varying ensemble size shows how the signal depends on the number of simulators: contrast AUC rises from $\aucKtwo$ at $K{=}2$ to $\aucKsix$ at $K{=}6$ (Figure~\ref{fig:robust}D), as more simulators estimate the simulator-specific spread more precisely.

\begin{table}[t]
\centering
\begin{minipage}{0.51\textwidth}\centering
  \caption{Estimand difficulty: invalidity and mean error by family
  ($n$ over resolved simulator--estimand pairs).}\label{tab:difficulty}
  \small\setlength{\tabcolsep}{4pt}
\begin{tabular}{l S[table-format=4.0] S[table-format=1.3] S[table-format=1.3]}
\toprule
{Estimand family} & {$n$} & {invalidity} & {mean $|\delta|$} \\
\midrule
direction & 3570 & 0.391 & 0.415 \\
ordinal contrast & 3251 & 0.507 & 0.444 \\
mean & 3824 & 0.651 & 0.496 \\
distribution & 4157 & 0.942 & 0.488 \\
\bottomrule
\end{tabular}

\end{minipage}\hfill
\begin{minipage}{0.47\textwidth}\centering
  \caption{Bias--variance decomposition per simulator on the same-unit families.}
  \label{tab:biasvar}
  \small\setlength{\tabcolsep}{4pt}
\begin{tabular}{l S[table-format=1.3] S[table-format=1.3] S[table-format=1.3] S[table-format=1.3]}
\toprule
{Simulator} & {RMSE} & {bias} & {std} & {v.frac} \\
\midrule
mistral-small & 0.540 & 0.515 & 0.161 & 0.089 \\
gpt-4o-mini & 0.589 & 0.586 & 0.055 & 0.009 \\
gpt-4o & 0.621 & 0.617 & 0.063 & 0.010 \\
qwen3-next-80b & 0.644 & 0.644 & 0.026 & 0.002 \\
mistral-large & 0.674 & 0.671 & 0.070 & 0.011 \\
gemma4-31b & 0.679 & 0.679 & 0.014 & 0.000 \\
\bottomrule
\end{tabular}

\end{minipage}
\end{table}

\paragraph{Prediction on test populations} Risk scores for test populations use simulator outputs, with any learned mapping fixed on calibration populations. Their predictive performance (Figure~\ref{fig:main}) supports transfer across countries within these surveys for the estimand families \mbox{where disagreement is informative}.

\section{Calibration and cost-based referral}
\label{app:decision}
The $80\%$ coverage used above is an illustrative operating point. To choose a threshold from assigned error and referral costs, we fit probabilities of invalidity for ordinal contrasts on calibration populations and evaluate them on separate test populations. This supplements the main response-proportion analysis, which uses disagreement directly and gives invalidity of $\propInvalidFullPct\%$ at full coverage and $\propInvalidEightyPct\%$ at $80\%$ coverage.

\begin{figure}[H]
\centering
\begin{subfigure}{0.44\textwidth}\centering
  \includegraphics[width=\linewidth]{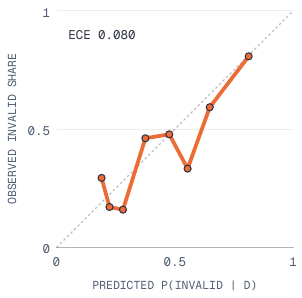}
  \caption{Reliability of $P(\mathrm{invalid}\mid D)$.}\label{fig:calib}
\end{subfigure}\hfill
\begin{subfigure}{0.46\textwidth}\centering
  \includegraphics[width=\linewidth]{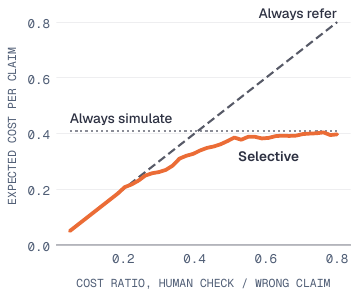}
  \caption{Expected cost vs.\ cost ratio.}\label{fig:cost}
\end{subfigure}
\caption{\textbf{From risk score to decision (contrast, test populations).}
(a) Probability calibration on test populations (ECE $\eceGap$).
(b) Expected cost under the assigned error and referral costs.}
\end{figure}

\paragraph{Probability calibration}
Specifically, a logistic model estimates $P(\mathrm{invalid}\mid D)$ on calibration populations. We bin its test predictions to compare predicted and observed invalidity. Figure~\ref{fig:calib} shows the reliability curve: predicted and observed invalidity track the diagonal, with an expected calibration error of $\eceGap$. These test-population probabilities provide the inputs to the referral decision below.

\paragraph{Referral under assigned costs}
Using these predicted probabilities, let $c_\mathrm{wrong}$ be the cost assigned to an invalid simulated estimate and $c_\mathrm{human}$ the cost assigned to referral. We assume human referral resolves the estimate. Retaining an estimate has expected cost $P(\mathrm{invalid}_j)\,c_\mathrm{wrong}$, while referral costs $c_\mathrm{human}$. Under this model, referral is optimal when $P(\mathrm{invalid}_j)>r=c_\mathrm{human}/c_\mathrm{wrong}$. Figure~\ref{fig:cost} compares this rule with retaining every estimate or referring every estimate. The maximum modeled saving is $\maxSavingPct\%$. At $r=0.3$, the rule retains $\coverageRthreePct\%$ of contrasts and saves $\savingRthreePct\%$. The savings are calculated from the assigned costs rather \mbox{than measured study expenditure}.

\section{Exploratory interactive analysis}
\label{app:interactive}
The survey analysis concerns one-shot responses. To examine whether its level--contrast ordering also appears in interactive data, we compare simulator outputs with recorded human donation decisions in a persuasion-dialogue corpus~\citep{wang2019persuasion}.

All simulators overstate engagement, and contrast error is lower than level error. The overstatement is sufficiently uniform that the disagreement predictor cannot be scored on these outcomes. The case therefore permits a level--contrast error comparison, while evaluating disagreement as a predictor requires an interactive corpus covering a wider range of errors.

\end{document}